\pdfoutput=1

\documentclass[11pt]{article}

\usepackage[]{ACL2023}

\usepackage{times}
\usepackage{latexsym}

\usepackage[T1]{fontenc}

\usepackage[utf8]{inputenc}

\usepackage{microtype}

\usepackage{inconsolata}

\usepackage{booktabs}
\usepackage{multirow}
\usepackage{multicol}
\usepackage{amsmath}
\usepackage{enumitem}
 \usepackage{graphicx}

\definecolor{wxjiao}{RGB}{18, 141, 21}

\title{ChatGPT or Grammarly? Evaluating ChatGPT on Grammatical Error Correction Benchmark}

\author{
Haoran Wu$^\dagger$ \quad Wenxuan Wang$^\dagger$  \quad Yuxuan Wan $^\dagger$\quad Wenxiang Jiao$^\ddagger$
\quad  Michael R. Lyu$^\dagger$\\
$^\dagger$Department of Computer Science and Engineering, The Chinese University of Hong Kong\\ { \normalsize \texttt{1155157061@link.cuhk.edu.hk} \quad \texttt{\{wxwang,yxwan9,lyu\}@cse.cuhk.edu.hk}} \\
$^\ddagger$Tencent AI Lab \\
{ \normalsize \texttt{joelwxjiao@tencent.com}}
}

\begin{document}

\maketitle

\begin{abstract}
ChatGPT is a cutting-edge artificial intelligence language model developed by OpenAI, which has attracted a lot of attention due to its surprisingly strong ability in answering follow-up questions. In this report, we aim to evaluate ChatGPT on the Grammatical Error Correction~(GEC) task, and compare it with commercial GEC product (e.g., Grammarly) and state-of-the-art models (e.g., GECToR). By testing on the CoNLL2014 benchmark dataset, we find that ChatGPT performs not as well as those baselines in terms of the automatic evaluation metrics (e.g., $F_{0.5}$ score), particularly on long sentences. We inspect the outputs and find that ChatGPT goes beyond one-by-one corrections. Specifically, it prefers to change the surface expression of certain phrases or sentence structure while maintaining grammatical correctness. Human evaluation quantitatively confirms this and suggests that ChatGPT produces less under-correction or mis-correction issues but more over-corrections. These results demonstrate that ChatGPT is severely under-estimated by the automatic evaluation metrics and could be a promising tool for GEC. 
\end{abstract}

\section{Introduction}

ChatGPT\footnote{\url{https://chat.openai.com/chat}}, the current ``super-star'' in artificial intelligence~(AI) area, has attracted millions of registered users within just a week since its launch by OpenAI.
One of the reasons for ChatGPT being so popular is its surprisingly strong performance on various natural language processing~(NLP) tasks~\cite{bang2023M3ChatGPT}, including question answering~\cite{Omar2023ChatGPTVT}, text summarization~\cite{Yang202ChatGPT4Summary}, machine translation~\cite{jiao2023ischatgpt}, logic reasoning~\cite{Frieder2023MathChatGPT}, code debugging~\cite{Xia2023ConversationalAP}, etc. There is also a trend of using ChatGPT as a writing assistant for text polishing.


Despite the widespread use of ChatGPT, it remains unclear to the NLP community that to what extent ChatGPT is capable of revising the text and correcting grammatical errors.
To fill this research gap, we empirically study the Grammatical Error Correction~(GEC) ability of ChatGPT by evaluating on the CoNLL2014 benchmark dataset~\cite{Ng2014TheCS}, and comparing its performance to Grammarly, a prevalent cloud-based English typing assistant with 30 million users daily~\cite{Grammarly_User} and GECToR~\cite{Omelianchuk2020GECToRG}, a state-of-the-art GEC model. 
With this study, we aim to answer a research question:
\begin{quote}
    Is ChatGPT a good tool for GEC?
\end{quote}
To the best of our knowledge, this is the first study on ChatGPT's ability in GEC.


We present the major insights gained from this evaluation as below:
\begin{itemize}
    \item ChatGPT performs worse than the baseline systems in terms of the automatic evaluation metrics (e.g., $F_{0.5}$ score), particularly on long sentences.
    \item ChatGPT goes beyond one-by-one corrections by introducing more changes to the surface expression of certain phrases or sentence structure while maintaining the grammatical correctness.
    \item Human evaluation quantitatively demonstrates that ChatGPT produces less under-correction or mis-correction issues but more over-corrections. 
\end{itemize}
Our evaluation indicates the limitation of relying solely on automatic evaluation metrics to assess the performance of GEC models
and suggests that ChatGPT is a promising tool for GEC.

\begin{table*}[t!]
\centering
    \begin{tabular}{l| cc}
    \toprule
    \bf Type & \multicolumn{1}{c}{\bf Error} & \multicolumn{1}{c}{\bf Correction} \\ 
    \midrule
    Preposition &  I sat in the talk & I sat in on the talk \\
    Morphology  &  dreamed &  dreamt \\
    Determiner &   I like the ice cream &  I like ice cream  \\
    Tense/Aspect   &    I like play basketball  &  I like playing basketball \\
    Syntax   &  I have not the book  &  I do not have the book \\
    Punctuation  &  We met they talked and left & We met, they talked and left \\
    \bottomrule
    \end{tabular}
    \caption{Different types of error in GEC.}
    \label{tab:type}
\end{table*}

\section{Background}

\subsection{ChatGPT}

ChatGPT is an intelligent chatbot powered by large language models developed by OpenAI. It has attracted great attention from industry, academia, and the general public due to its strong ability in answering various follow-up questions, correcting inappropriate questions~\cite{Zhong2023CanCU}, and even refusing illegal questions.
While the technical details of ChatGPT have not been released systematically, it is known to be built upon InstructGPT~\cite{ouyang2022InstructGPT} which is trained using instruction tuning~\cite{Wei2021FinetunedLM} and reinforcement learning from human feedback~\cite[RLHF,][]{Christiano2017DeepRL}.




\subsection{Grammatical Error Correction}
Grammatical Error Correction (GEC) is a task of correcting different kinds of errors in text such as spelling, punctuation, grammatical, and word choice errors~\cite{Ruder_NLP-progress_2022}.
It is highly demanded as writing plays an important role in academics, work, and daily life.
Table~\ref{tab:type} presents the illustration of different grammatical errors borrowed from \newcite{Bryant2022GrammaticalEC} in a comprehensive survey on grammatical error correction. In general, grammatical errors can be roughly classified into three categories: omission errors, such as "on" in the first example; replacement errors, such as "dreamed" for "dreamt" in the second example; and insertion errors, such as "the" in the third example.

To evaluate the performance of GEC, researchers have built various benchmark datasets, which include but are not limited to:
\begin{itemize}
    \item \textbf{CoNLL-2014}: Given the short English texts written by non-native speakers, the task requires a participating system to correct all errors present in each text.
    \item \textbf{BEA-2019}: It is similar to CoNLL-2014 but introduces a new dataset, namely, the Write\&Improve+LOCNESS corpus, which represents a wider range of native and learner English levels and abilities \cite{Bryant2019TheBS}.
    \item \textbf{JFLEG}: It represents a broad range of language proficiency levels and uses holistic fluency edits to not only correct grammatical errors but also make the original text more native sounding~\cite{Tetreault2017JFLEGAF}.
\end{itemize}

\begin{table}[t!]
\centering
    \begin{tabular}{l ccc}
    \toprule
    \bf System & \bf Precision & \bf Recall & \bf $F_{0.5}$\\ 
    \midrule
    GECToR & \bf 71.2 & 38.4 & 60.8 \\
    Grammarly & 67.3 & 51.1 & \bf 63.3 \\
    \hline
    ChatGPT & 51.2 & \bf 62.8 & 53.1\\
    \bottomrule
    \end{tabular}
    \caption{GEC performance of GECToR, Grammarly, and ChatGPT.}
    \label{tab:table1}
\end{table}

\section{ChatGPT for GEC}

\subsection{Experimental Setup}
\paragraph{Dataset.}
We evaluate the ability of ChatGPT in grammatical error correction on the CoNLL2014 task~\cite{Ng2014TheCS} dataset. The dataset is composed by short paragraphs that are written by non-native speakers of English, accompanied with the corresponding annotations on the grammatical errors.
We pulled 100 sentences from the official-combined test set in the alternate folder of the dataset sequentially.

\begin{table*}[t!]
\setlength{\tabcolsep}{3pt}
\centering
    \begin{tabular}{l ccc ccc ccc}
    \toprule
    \multirow{2}{*}{\bf System} & \multicolumn{3}{c}{ \bf Short} & \multicolumn{3}{c}{ \bf Medium} & \multicolumn{3}{c}{\bf Long}\\
    \cmidrule(lr){2-4} \cmidrule(lr){5-7} \cmidrule(lr){8-10}
     & Precision & Recall & $F_{0.5}$ & Precision & Recall & $F_{0.5}$ & Precision & Recall & $F_{0.5}$\\
    \midrule
    GECToR & 76.9 & 38.5 & 64.1  & 68.8 & 37.5 & 58.9 & 71.8 & 38.9 & 61.5\\
    Grammarly & 62.5 & 60.6 & 62.1  & 68.9 & 56.0 & 65.9 & 67.3 & 45.3 & 61.4\\
    \hline
    ChatGPT & 58.5 & 66.7 & 60.0  & 48.7 & 60.7 & 50.7 & 51.0 & 62.8 & 53.0 \\
    \bottomrule
    \end{tabular}
    \caption{GEC performance with respect to sentence length.}
    \label{tab:table3}
\end{table*}

\paragraph{Evaluation Metric.} 
To evaluate the performance of GEC, we adopt three metrics that are widely used in literature, namely, Precision, Recall, and $F_{0.5}$ score. Among them, $F_{0.5}$ score combines both Precision and Recall, where Precision is assigned a higher weight~\cite{enwiki:1139808388}. Specifically, the three metrics are expressed as:
\begin{align}
    &\mathrm{Precision} = \frac{TP}{TP + FP}, \\
    &\mathrm{Recall} = \frac{TP}{TP + FN}, \\
    &\mathrm{F_{0.5}} = \frac{\rm 1.25\times Precision\times Recall}{\rm 0.25\times Precision + Recall},
\end{align}
where $TP$, $FP$ and $FN$ represent the true positives, false positives and false negatives of the predictions, respectively.
We use the scoring program provided by CoNLL2014 official but adapt it to be compatible with the latest Python environment.

\paragraph{Baselines.}
In this report, we perform the GEC task on three systems, including:
\begin{itemize}[leftmargin=10pt]
    \item \textbf{ChatGPT}: We query ChatGPT manually rather than using some API due to the instability of ChatGPT. For example, when a query sentence resembles a question or demand, ChatGPT may stop the process of GEC but respond to the ``demand'' instead. After a few trials, we find a prompt that works well for ChatGPT:
    \begin{quote}
        \tt Do grammatical error correction on all the following sentences I type in the conversation.
    \end{quote}
    We query ChatGPT with this prompt for each test sample.
    \item \textbf{Grammarly}: Grammarly is a prevalent cloud-based English typing assistant. It reviews spelling, grammar, punctuation, clarity, engagement, and delivery mistakes in English texts, detects plagiarism and suggests replacements for the identified errors \cite{enwiki:1139043929}. As stated by Grammarly, every day, 30 million people and 50,000 teams around the world use Grammarly with their writing \cite{Grammarly_User}.
    When querying Grammarly, we open a text file and paste all the test samples into separate paragraphs. We enable all the grammar correction in the setting and only ask it to correct the ones with correctness problems (\textcolor{red}{\underline{red underline}}), while leaving the clarity (\textcolor{blue}{\underline{blue underline}}), engagement (\textcolor{green}{\underline{green underline}}) and delivery (\textcolor{purple}{\underline{purple underline}}) unchanged. We iterate this process several times until there is no error detected by Grammarly.
    \item \textbf{GECToR}: Besides Grammarly, we also compare ChatGPT with GECToR~\cite{Omelianchuk2020GECToRG}, a state-of-the-art model on GEC in research, which also exhibits good performance on the CoNLL2014 task. We adopt the implementation based on the pre-trained RoBERTa model. 
\end{itemize}

\subsection{Results and Analysis}
\paragraph{Overall Performance.}
Table~\ref{tab:table1} presents the overall performance of the three systems.
As seen, ChatGPT obtains the highest recall value, GECToR obtains the highest precision value, while Grammarly achieves a better balance between the two metrics and results in the highest $F_{0.5}$ score.
These results suggest that ChatGPT tends to correct as many errors as possible, which may lead to more overcorrections. Instead, GECToR corrects only those it is confident about, which leaves many errors uncorrected. Grammarly combines the advantages of both such that it performs more stably.

\begin{table*}[t!]
\centering
    \begin{tabular}{l p{12cm}}
    \toprule
    \bf System & \multicolumn{1}{c}{\bf Sentence} \\
    \midrule
    Source & For \textcolor{red}{an} example , if exercising is helpful for family potential disease , we can always look for more chances for the family to go exercise .\\
    Reference & For example , if exercising \textcolor{red}{(OR exercise)} is helpful for \textcolor{red}{a potential family} disease , we can always look for more chances for the family to \textcolor{red}{do} exercise .\\
    \hline
    GECToR & For example , if exercising is helpful for family potential disease , we can always look for more chances for the family to go exercise .\\
    Grammarly & For example , if exercising is helpful for \textcolor{red}{a} family \textcolor{red}{'s} potential disease , we can always look for more chances for the family to go exercise .\\
    ChatGPT & For example , if exercise is helpful \textcolor{red}{in preventing} potential family \textcolor{red}{diseases} , we can always look for more \textcolor{red}{opportunities} for the family to exercise .\\
    \bottomrule
    \end{tabular}
    \caption{Comparison of the outputs from different GEC systems.}
    \label{tab:system_example}
\end{table*}

\paragraph{ChatGPT Performs Worse on Long Sentences?}
To understand which kind of sentences ChatGPT are good at, we divide the 100 test sentences into three equally sized categories, namely, Short, Medium and Long.
Table~\ref{tab:table3} shows the results with respect to sentence length. As seen, the gap between ChatGPT and Grammarly is significantly bridged on short sentences. In contrast, ChatGPT performs much worse on those longer sentences, at least in terms of the existing evaluation metrics.

\begin{table}[]
\centering
    \begin{tabular}{l ccc}
    \toprule
    \bf System & \bf Precision & \bf Recall & \bf $F_{0.5}$ \\ 
    \midrule
    GECToR &  71.2 & 38.4 & 60.8 \\
    ~~~ + Grammarly & -5.9 &  +16.5 &  +2.1 \\
    \hline
    ChatGPT & 51.2 & 62.8 & 53.1 \\
    ~~~ + Grammarly &  +0.4 &  +0.8 & +0.5 \\
    \bottomrule
    \end{tabular}
    \caption{GEC performance with Grammarly for further correction.}
    \label{tab:further-GEC-by-grammarly}
\end{table}

\paragraph{ChatGPT Goes Beyond One-by-One Corrections.}
We inspect the output of the three systems, especially those for long sentences, and find that ChatGPT is not limited to correcting the errors in the one-by-one fashion. Instead, it is more willing to change the superficial expression of some phrases or the sentence structure. For example, in Table~\ref{tab:system_example}, GECToR and Grammarly make minor changes to the source sentence (i.e., ``an example'' to ``example'', ``family potential disease'' to ``a family 's potential disease''), while ChatGPT modifies the sentence structure (i.e., ``for family potential disease'' to ``in preventing potential family diseases'') and word choice (i.e., ``chances'' to ``opportunities''). It indicates that the outputs by ChatGPT maintain the grammatical correctness, although they do not follow the original expression of the source sentences.

To validate our hypothesis, we let Grammarly to further correct the grammatical errors in the outputs of GECToR and ChatGPT. Table~\ref{tab:further-GEC-by-grammarly} lists the results. We can observe that Grammarly introduces a negligible improvement to the output of ChatGPT, demonstrating that ChatGPT indeed generates correct sentences. On the contrary, Grammarly further improves the performance of GECToR noticeably (i.e., +2.1 $F_{0.5}$, +16.5 Recall), suggesting that there are still many errors in the output of GECToR.

\paragraph{Human Evaluation.}

We conduct a human evaluation to further demonstrate the potential of ChatGPT for the GEC task. Specifically, we follow~\newcite{wang2022understanding} to manually annotate the issues in the outputs of the three systems, including 1) Under-correction, which is the grammatical errors that are not found; 2) Mis-correction, which is the grammatical errors that are found but modified incorrectly; it can be either grammatically incorrect or semantically incorrect; 3) Over-correction, which is the other modifications beyond the changes in the reference. 
We sample 20 sentences out of the 100 test sentences and ask two annotators to identify the issues.
Table~\ref{tab:human-eval} shows the results. Obviously, ChatGPT has the least number of under-corrections among the three systems and fewer number of mis-corrections compared with GECToR, which suggests its great potential in grammatical error correction. Meanwhile, ChatGPT produces more over-corrections, which may come from the diverse generation ability as a large language model. While this usually leads to a lower $F_{0.5}$ score, it also allows more flexible language expressions in GEC.

\begin{table}[]
\centering
    \begin{tabular}{l ccc}
    \toprule
\bf System & \bf \#Under & \bf \#Mis & \bf \#Over \\
    \midrule
    GECToR & 13 & 4 & 0\\
    Grammarly & 14 & 0 & 1 \\
        \hline
    ChatGPT &  3 & 3 & 30\\
    \bottomrule
    \end{tabular}
    \caption{Number of under-correction~(Under), mis-correction (Mis) and over-correction~(Over) produced by different GEC systems.}
    \label{tab:human-eval}
\end{table}

\paragraph{Discussions.}
We have checked the outputs corresponding to the results of Table~\ref{tab:further-GEC-by-grammarly}, and observed different behaviors of ChatGPT and Grammarly. The slight improvement (i.e., +0.5 $F_{0.5}$) by Grammarly mainly comes from punctuation problems. ChatGPT is not sensitive to punctuation problems but Grammarly is, though the modifications are not always correct. For example, when we manually undo the corrections on punctuation, the $F_{0.5}$ score increases by +0.0015. 
Other than punctuation problems, Grammarly also corrects a few grammatical errors on articles, prepositions, and plurals. However, these corrections usually require Grammarly to repeat the process twice. Take the following sentence as an example,
\begin{quote}
    \tt ... constructs of the family and kinship {\color{red}are a social construct}, ...
\end{quote}
Grammarly first changes it to
\begin{quote}
    \tt ... constructs of the family and kinship {\color{red}are a social constructs}, ...
\end{quote}
Then, changes it to
\begin{quote}
    \tt ... constructs of the family and kinship {\color{red}are social constructs}, ...
\end{quote}
Nonetheless, it does correct some errors that ChatGPT fails to correct.


\section{Conclusion}
This paper evaluates ChatGPT on the task of Grammatical Error Correction (GEC). By testing on the CoNLL2014 benchmark dataset, we find that ChatGPT performs worse than a commercial product Grammarly and a state-of-the-art model GECToR in terms of automatic evaluation metrics.
By examining the outputs, we find that ChatGPT displays a unique ability to go beyond one-by-one corrections by changing surface expressions and sentence structure while maintaining grammatical correctness. 
Human evaluation results confirm this finding and reveals that ChatGPT produces fewer under-correction or mis-correction issues but more over-corrections. These results demonstrate the limitation of relying solely on automatic evaluation metrics to assess the performance of GEC models and suggest that ChatGPT has the potential to be a valuable tool for GEC.

\section*{Limitations and Future Works}

There are several limitations in this version, which we leave for future work:
\begin{itemize}[leftmargin=10pt]
    \item \textbf{More Datasets}: In this version, we only use the CoNLL-2014 test set and only randomly select 100 sentences to conduct the evaluation. In our future work, we will conduct experiments on more datasets.
    \item \textbf{More Prompt and In-context Learning}: In this version, we only use one prompt to query ChatGPT and do not utilize the advanced technology from the in-context learning field, such as providing demonstration examples~\cite{brown2020gpt3} or providing chain-of-thought~\cite{Wei2022ChainOT}, which may under-estimate the full potential of ChatGPT. In our future work, we will explore the in-context learning methods for GEC to improve its performance.
    \item \textbf{More Evaluation Metrics}: In this version, we only adopt Precision, Recall and $F_{0.5}$ as evaluation metrics. In our future work, we will utilize more metrics, such as pretraining-based metrics~\cite{Gong2022RevisitingGE} to evaluate the performance comprehensively.
\end{itemize}

\bibliography{anthology,custom}
\bibliographystyle{acl_natbib}

\end{document}